%% file: main.tex
\documentclass[letterpaper, 10 pt, conference]{ieeeconf}  
\IEEEoverridecommandlockouts                              
\usepackage{graphicx}
\usepackage{epsfig} 
\usepackage{mathptmx}
\usepackage{mathtools}
\usepackage{amsmath} 
\usepackage{bm}

\usepackage{amssymb}  
\usepackage{amsthm}
\usepackage{siunitx}
\usepackage{wasysym}
\usepackage{wasysym}
\usepackage{xcolor} 
\usepackage[mathcal]{euscript}
\usepackage{bbm}
\usepackage{color}
\usepackage{lipsum}
\usepackage{booktabs}
\usepackage{hyperref}
\hypersetup{colorlinks=true,linkcolor=black,anchorcolor=black,citecolor=black,filecolor=black,menucolor=black,runcolor=black,urlcolor=black}

\usepackage[ruled,vlined,linesnumbered]{algorithm2e}
\usepackage{etoolbox}
\usepackage{bigfoot}
\usepackage{lipsum}

\newcommand\blfootnote[1]{%
  \begingroup
  \renewcommand\thefootnote{}\footnote{#1}%
  \addtocounter{footnote}{-1}%
  \endgroup
}
\usepackage{multirow}
\usepackage{bbm}
\usepackage{outlines}
\let\labelindent\relax
\usepackage{enumitem}
\setenumerate[2]{label=\alph*.}
\setenumerate[3]{label=\roman*.}
\usepackage[capitalise]{cleveref}
\crefname{equation}{Eq.}{Eqs.}
\crefname{figure}{Fig.}{Figs.}
\crefname{tabular}{Tab.}{Tabs.}
\crefname{section}{Sec.}{Secs.}

\makeatletter
\patchcmd{\@makecaption}
  {\scshape}
  {}
  {}
  {}
\makeatother

\usepackage[
    style=ieee,
    doi=false,
    isbn=false,
    url=false,
    eprint=false,
    backend=biber,
    natbib=true
    ]{biblatex}

\usepackage{caption}
\bibliography{references,zotero}

\usepackage{listings}
\usepackage{tcolorbox}
\tcbuselibrary{listings}
\usepackage{booktabs}

\newtcblisting{mycodelisting}{%
  listing only,
  listing engine=listings,
  colback=gray!5,
  colframe=gray!80,
  boxrule=0.4pt,
  arc=2pt,
  outer arc=2pt,
  top=3pt,
  bottom=3pt,
  left=6pt,
  right=6pt,
  fonttitle=\bfseries,
  title=,
  listing options={
    language=Python,
    basicstyle=\ttfamily\small,
    breaklines=true,
    tabsize=4,
    frame=none
  }
}

\input{macros}

\title{\LARGE \bf
Seeing, Saying, Solving: \\ An LLM‑to‑TL Framework for Cooperative Robots}
\author{
Dan BW Choe\tss{1}, Sundhar Vinodh Sangeetha\tss{2}, Steven Emanuel\tss{3}, \\Chih-Yuan Chiu\tss{1}, Samuel Coogan\tss{1} and Shreyas Kousik\tss{3}
\thanks{
All authors are with the Georgia Institute of Technology, Atlanta, GA, USA.
\tss{1} School of Electrical and Computer Engineering.
\tss{2} School of Aerospace Engineering.
\tss{3} School of Mechanical Engineering.
Corresponding author: \texttt{bchoe7@gatech.edu}
}
}

\begin{document}
\makeatletter
\let\@oldmaketitle\@maketitle
\renewcommand{\@maketitle}{\@oldmaketitle
    \centering
    \captionsetup{type=figure}
    \setcounter{figure}{0}

    \vspace{5pt}
    \includegraphics[width=\linewidth]{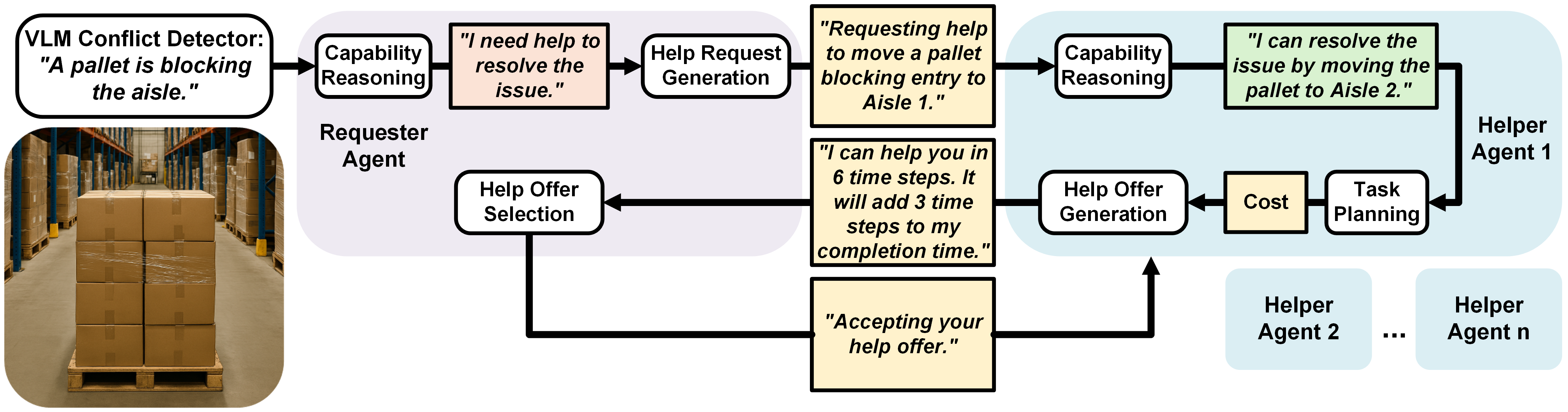}
    \captionof{figure}{Overview of the proposed framework. 
    The requester agent encounters a conflict and broadcasts a help request to other agents if needed.
    A potential helper agent receives the message and reasons whether it has the required capabilities to assist. 
    If able to assist, the potential helper agent computes the cost of helping and includes it in a help offer sent to the requester.
    The requester selects the help offer with the lowest cost and sends a confirmation message to the selected helper.
    }
    \label{fig:overall_framework}
    \vspace{-5pt}
}
\makeatother

\maketitle
\pagestyle{plain} 

\input{sections/00_abstract}

\input{sections/01_intro}
\input{sections/03_problem}
\input{sections/04_method}
\input{sections/05_experiments}
\input{sections/06_conclusion}


\renewcommand{\bibfont}{\normalfont\footnotesize}
{\renewcommand{\markboth}[2]{}
\printbibliography}
\onecolumn

\begin{appendices}
\crefalias{section}{appendix}
\crefalias{subsection}{appendix}
\input{sections/08_appendix}

\end{appendices}
\end{document}

%% file: macros.tex
\newtheorem*{problem*}{Problem}

\providecommand{\R}{\ensuremath \mathbb{R}}
\providecommand{\N}{\ensuremath \mathbb{N}}
\providecommand{\Z}{\ensuremath \mathbb{Z}}
\newcommand{\NL}{\mathbb{L}}

\newcommand{\regtext}[1]{\mathrm{\textnormal{#1}}}

\newcommand{\tss}[1]{\textsuperscript{#1}}

\newcommand{\lbl}[1]{^{\regtext{#1}}}

\newcommand{\st}{\,\regtext{s.t.}\,}

\DeclareMathOperator*{\argmin}{arg\,min}
\newcommand{\given}{\,\mid\,}

\newcommand{\proposition}{\varphi}
\newcommand{\trajectory}{X}
\newcommand{\always}{\Box}
\newcommand{\eventually}{\Diamond}

\newcommand{\gridworld}{W}
\newcommand{\freecells}{F}
\newcommand{\position}{x}
\newcommand{\duration}{\tau}
\newcommand{\msg}{m}
\newcommand{\allrobots}{\mathcal{R}}
\newcommand{\nrobots}{n}
\newcommand{\makespan}{T}
\newcommand{\confmsg}{\msg\conf}
\newcommand{\confposition}{\position\conf}
\newcommand{\confduration}{\duration\conf}
\newcommand{\allhelpers}{\mathcal{H}}
\newcommand{\capabilities}{c}

\newcommand{\distance}{D}
\DeclareMathOperator{\reason}{NLR}
\newcommand{\prompt}{p}
\DeclareMathOperator{\bnf}{BNF}
\newcommand{\timehorizon}{H}

\newcommand{\glob}{\lbl{g}}
\newcommand{\help}{\lbl{h}}
\newcommand{\conf}{\lbl{c}}
\newcommand{\offr}{\lbl{o}}
\newcommand{\req}{\lbl{r}}
\newcommand{\selection}{\lbl{s}}
\newcommand{\updated}{\lbl{new}}
\newcommand{\opt}{^\star}
\newcommand{\orig}{\lbl{orig}}


%% file: sections/00_abstract.tex
\begin{abstract}
Increased robot deployment, such as in warehousing, has revealed a need for seamless collaboration among heterogeneous robot teams to resolve unforeseen conflicts.
To address this challenge, we propose a novel, decentralized framework for robots to request and provide help.
The framework begins with robots detecting conflicts using a Vision Language Model (VLM), then reasoning over whether help is needed.
If so, it crafts and broadcasts a natural language (NL) help request using a Large Language Model (LLM).
Potential helper robots reason over the request and offer help (if able), plus information about impact to their current tasks.
Helper reasoning is implemented via an LLM grounded in Signal Temporal Logic (STL) using a Backus–Naur Form (BNF) grammar to guarantee syntactically valid NL to STL translations, which are then solved as a Mixed Integer Linear Program (MILP).
Finally, the requester robot chooses a helper by reasoning over impact on the overall system.
We evaluate our system via experiments considering different strategies for how to choose a helper, and find that a requester robot can minimize overall time impact on the system by considering multiple help offers versus simple heuristics (e.g. selecting the nearest robot to help).
\end{abstract}

%% file: sections/01_intro.tex
\section{Introduction} \label{sec:intro}


Robots ranging from drones to autonomous forklifts are increasingly common in modern warehouses, where semi-structured environments enable large-scale deployment of heterogeneous robot teams.
However, the dynamic and shared nature of these spaces often leads to physical and semantic conflicts that are difficult to resolve without human intervention.

Large vision-language models (VLMs) offer a potential solution via their strengths in scene understanding and anomaly detection \cite{sinha2024real} that can enable conflict resolution in multi-robot settings \cite{kato2024design}. 
However, the plans generated by VLMs cannot guarantee safety, a primary concern in shared human-robot environments. 
In addition, warehouse robots frequently operate under strict time constraints, such as in order fulfillment scenarios where shipments must be staged at outbound docks by designated pickup times.
Without guarantees on the safety and feasibility of proposed conflict recovery plans, VLMs alone are insufficient for ensuring reliable multi-agent conflict resolution in these time-sensitive and safety-critical environments. 

One path forward to solving this challenge is to apply formal methods, such as the verification of Signal Temporal Logic (STL) and Linear Temporal Logic (LTL) specifications.
These can guarantee that a robot's plan satisfies constraints and have seen success in multi-robot coordination and reconfiguration tasks \cite{guo2015multi}.
Recent work has also explored their integration with large language models (LLMs) to enforce safety constraints during task planning \cite{yang2024plug}.
However, it remains open how to integrate these formal methods approaches with foundation model reasoning in a way that enables seamless, safe multi-agent coordination.
Note, \cref{appx: related work} provides a more detailed overview of related work.


\subsubsection*{Related Work}
This work lies at the intersection of foundation models and formal methods for multi-agent robot task planning.
Closest to our work, \cite{kato2024design} uses foundation models to detect, describe, and reason over conflicts and robot capabilities, but without formal guarantees.
Similarly, in formal methods, \cite{liu2024lang2ltl}, uses a VLM to ground spatio-temporal navigation commands in unseen environments for a single agent.
In contrast to the previous works, we propose a novel multi-agent framework that integrates STL to augment an LLM agent with spatial and temporal reasoning capabilities.

\subsubsection*{Contributions}
As a step towards addressing the above research question, we present a framework that enables heterogeneous warehouse robots, each subject to local task specifications, to request and offer assistance in natural language, while ensuring safety and feasibility of the resulting recovery plan through formal verification.
In particular, we make the following contributions:
\begin{enumerate}
    \item We propose a method to transform natural language specifications into temporal logic specifications with a guarantee on validity (\cref{subsec: method: help request generation,subsec: method: translating proposals to stl}).

    \item We augment LLM agents with spatial and temporal reasoning capabilities, therefore making progress towards realizing guarantees from formal methods (\cref{subsec: method: optimal path via milp}).

    \item We present an experimental implementation of the proposed framework in simulation
    (\cref{sec: experiments}).
\end{enumerate}

%% file: sections/03_problem.tex
\section{Preliminaries and Problem Statement}\label{sec: problem statement}

Consider robots operating in parallel on a variety of tasks; in this work we use warehousing as our canonical example.
\textbf{If a robot detects a conflict that it cannot resolve on its own, how can it request and receive help from a fellow robot while ensuring safety and minimizing impact on the overall system?}
We now formalize these notions.

\subsubsection*{Notation}
The reals are $\R$, natural numbers are $\N$, and integers are $\Z$.
To allow mathematically describing functions operating on natural language, we denote $\NL$ as a set of all natural language utterances.
Subscripts are indices and superscripts are labels.
We use $\gets$ to denote the output of NL reasoning.

\subsubsection*{World and Robots}
We consider $\nrobots$ robots operating in a discrete time grid world, so robot $i$ is at $\position_i(t) \in \gridworld \subset \Z^2$ at time $t = 0,1,\cdots,\timehorizon$, where $\timehorizon \in \N$ is a finite time horizon.
Each cell is either free space or a static obstacle, a common setup in modeling MRS \cite{stern2019multi}; we denote all free cells as  $\freecells \subseteq \gridworld$.
Multiple robots can occupy a free cell concurrently ($\position_i(t) = \position_j(t) \in \freecells$ is allowed); we assume local coordination and collision avoidance are handled by lower-level planning and control.
Finally, each robot has an NL representation of its own capabilities $\capabilities_i \in \NL$ (e.g., ``can lift pallets'' or ``max speed $1~\si{m/s}$'')

\subsubsection*{Tasks}
Each robot has assigned tasks represented as an STL specification $\proposition_i$ and path $\trajectory_i$ for robot $i$.
All robots must also obey a global \blfootnote{Code available at: \url{https://github.com/dchoe1122/seeing_saying_solving}
}specification\footnote{\label{note2 }An example can be found in \cref{appx: global stl spec} (e.g., actuation limits and obstacles)} $\proposition\glob$.
We assume each robot's tasks are feasible and expressed in valid STL.
For any STL expression $\proposition$, we define $\makespan(\proposition) \in \N$ as the makespan, or minimum duration required to satisfy $\proposition$.

\subsubsection*{Communications}
Robots can send and receive natural language messages $\msg \in \NL$ instantaneously and error-free.
Denote $\msg_{i\to j}$ as a message from robot $i$ to robot $j$, and a broadcast message as $\msg_{i\to\allrobots}$, where $\allrobots = \{0,1,\cdots,\nrobots\}$ denotes all robots.

\subsubsection*{Conflicts}
Each robot can detect a conflict (e.g., using VLM conflict detector as in \cref{fig:overall_framework}), which we denote as a tuple $(\confmsg,\confposition,\confduration) \in \NL \times \gridworld\times \N$ containing a natural language description, a location, and a duration required to resolve it.
We assume $\confposition$ is resolved when a helper robot enters the same cell as the requester for at least $\confduration$ time steps.

\begin{problem*}
Suppose robot $i$ detects a conflict $(\confmsg,\confposition,\confduration)$.
Without using centralized task assignment, we seek to identify another robot $j\in \allhelpers_i$, where $\allhelpers_i = \allrobots \setminus \{i\}$ is the set of candidate helper robots, and plan its motion such that it resolves the conflict while minimizing the increase in the total makespan across all robots.
That is, create $\proposition_j$ such that $\position_j(t) = \confposition$ for all $t' \in [t,t+\confduration]$ while minimizing $\sum_{i \in \allrobots} \makespan(\proposition_i)$.    
\end{problem*}

%% file: sections/04_method.tex
\section{Proposed Method} \label{sec: method}


In our framework, a robot needing help (\textit{requester}) broadcasts a help request to the multi-robot system where each robot evaluates its capabilities and availability.
Then, any robot able to assist (\textit{helper}) formulates a help offer.
Prior to sending back the offer to the \textit{requester}, each \textit{helper} translates the help offer to STL and solves for the optimal path plan that minimizes total time impact to the system.
Once solved, the \textit{helper} responds with the help offer along with the projected time impact to the system.
Finally, the \textit{requester} chooses the \textit{helper} with the lowest impact by a help confirmation message.
\Cref{fig:overall_framework} illustrates our framework.

To proceed, \Cref{subsec: method: help request generation} covers NL processing, \Cref{subsec: method: translating proposals to stl} explains our NL to STL conversion, and \Cref{subsec: method: optimal path via milp} details solving for optimal paths.

\subsection{Generating Help Requests, Offers, and Confirmations}\label{subsec: method: help request generation}

We implement a multi-robot communication protocol for collaborative conflict resolution using Large Language Models.
This protocol involves three capabilities: send, receive and broadcast.
We define three types of messages: help requests, help offers, and help confirmations\footnote{An example help request is in \Cref{appx:example_request} and help offer in \Cref{appx:example_proposal}.}.

Help requests are broadcast messages sent by a robot facing a conflict after determining that it requires help, by reasoning over the conflict $(\confmsg, \confposition, \confduration)$ and its own capabilities $\capabilities_i$, conditioned on a prompt $\prompt\conf \in \NL$:
\begin{align}
    \msg_{i\to\allhelpers_i}\req \gets \reason(\confmsg, \confposition, \confduration, \capabilities_i \given \prompt\conf),
\end{align}
where $\reason$ abstractly represents NL reasoning, implemented with foundation models.
Help requests describe the scene, location, what is required for the conflict to be resolved, and why the robot cannot resolve the conflict.
To ensure conflict location and requester capabilities are included in help requests, we use constrained generation \cite{beurer2023prompting}.

Help offers are sent by potential helpers in response to a help request. 
To generate help offers, each robot $j$ reasons over its help request, its own location, and its capabilities:
\begin{align}
    \msg_{j \to i}\offr \gets \reason(\msg_{i\to\allhelpers_i}\req,\position_j,\capabilities_j \given \prompt\help).
\end{align}
We again use constrained generation to ensure help offers include information about capabilities.

In addition, the help offer includes the duration it will take robot $j$ to provide help $\duration\help_j$ (i.e., how long does robot $i$ need to wait?), and the additional time to help relative to completing its original tasks $\duration\updated_j$ (i.e., how much is robot $j$ impacted?).
We compute these durations below in \Cref{subsec: method: optimal path via milp}.

Finally, the requester confirms the lowest-cost help request.
That is, it finds $j\opt = \argmin_j (\duration\help_j + \duration\updated_j)$, then sends
$\msg_{i\to j\opt}\selection = \regtext{``accept''}$
and $\msg_{i\to j}\selection = \regtext{``reject''}$ for all $j \neq j\opt$.

\subsection{Translating Help Proposals to STL Specifications}\label{subsec: method: translating proposals to stl}

We draw from a body of recent work on the transformation of natural language tasks and specifications into temporal logic formulas using large language models \cite{chen2023nl2tl, fuggitti2023nl2ltl, mavrogiannis2024cook2ltl}.
We enhance these natural language to temporal logic methods by defining a Backus-Naur form (BNF) grammar, a notation system for defining formal languages that can be used to constrain the output of LLMs \cite{wang2023grammar}.
Specifically, defining a BNF grammar for STL involves defining unary and binary temporal and boolean operators, allowed predicates, and text representations of temporal logic relations.
These enforce the syntactic validity of generated temporal logic formulas, allowing us to directly feed LLM output into an STL solver\footnote{\label{encoding stl}Details on how we encode STL specifications can found in the appendix.}.
Ultimately, we generate an STL specification by reasoning over the help offer message and conflict location, conditioned on a prompt $\prompt\lbl{STL}$ and subject to the BNF grammar:
\begin{align}\label{eq: help offer to stl}
    \proposition_j\help \gets \reason(\msg_{j\to i}\offr,\position\conf \given \prompt\lbl{STL}, \bnf).
\end{align}


\subsection{Solving for Optimal Robot Paths}\label{subsec: method: optimal path via milp}


Suppose each candidate helper robot $j \in \allhelpers_i$ has an original task specification $\proposition\orig_j$ and has found a corresponding optimal path $\trajectory_j\orig = \{\position_j(t) \given t = 0,\cdots,\makespan(\proposition\orig_j)\}$ by minimizing Manhattan distance\footnote{\label{note1}We use Manhattan distance to capturing the number of movements in a  grid world and improve computation time over solving a Mixed Integer Quadratic Program (MIQP) with Euclidean Distances.} $\distance(\trajectory\orig_j)$ and makespan $\makespan(\proposition\orig_j)$ while always obeying global specifications:
\begin{subequations}\label{prog: original path milp}
\begin{align}
    \trajectory_j\orig = \argmin_\trajectory
        \quad & \distance(\trajectory) + \makespan(\proposition\orig_j) \\
        \st\qquad & \eventually\proposition\orig_j \land \always\proposition\glob.
\end{align}
\end{subequations}
Once the help offer $\msg\offr_{j\to i}$ is translated into $\proposition\help_j$ as in \eqref{eq: help offer to stl}, robot $j$ computes an updated path that both helps and completes the original tasks:
\begin{subequations}\label{prog: updated path milp}
\begin{align}
    \trajectory_j\updated = \argmin_\trajectory
        \quad & \distance(\trajectory) + \makespan(\proposition\updated_j) + \makespan(\proposition\help_j) \\
        \st\qquad & \proposition\updated_j = \eventually\proposition_j\help \land \eventually\proposition\orig_j \land \always\proposition\glob.
\end{align}
\end{subequations}
Importantly, we include $\makespan(\proposition\help_j)$ in the cost to let the helper robot minimize how long the requester must wait.

We formulate \eqref{prog: original path milp} and \eqref{prog: updated path milp} as Mixed Integer Linear Programs (MILPs) by defining predicates in the STL formula as binary variables over discrete time steps, and solve them with Gurobi \cite{gurobi}.

Finally, when crafting a help offer, robot $j$ reports the total time it will take to help \textit{and} the additional time to help relative to its original tasks:
\begin{align}
    \duration\help_j &= \makespan(\proposition\help_j) \\
    \duration\updated_j &= \makespan(\proposition\updated_j) - \makespan(\proposition\orig_j).
\end{align}






\begin{figure}
    \centering
    \includegraphics[width=6.5cm]{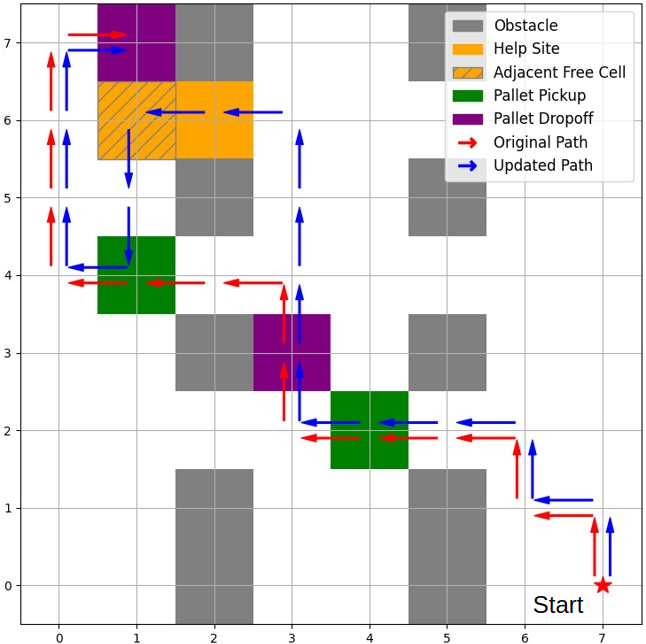}
    \caption{An example of a reconfigured path.
    The help site is reached in 11 time-steps, extending the original path by 4 time-steps for a total cost $\duration\help_j + \duration\updated_j = 15$ time-steps.}
    \vspace{-15pt}
    \label{fig:path}
\end{figure}

%% file: sections/05_experiments.tex
\section{Experiments} \label{sec: experiments}





\subsection{Experiment 1: Natural Language to STL}


In this experiment, we evaluate our method for transforming natural language (NL) to temporal logic (TL) using BNF constrained generation.
We implement our method with Gemma 3 27B LLM using the \texttt{llama.cpp} library for BNF constrained generation. The BNF grammar is included in the LLM prompt and as a constraint\footnote{An example BNF grammar can be found in \Cref{appx:bnf_example}.}.

\subsubsection*{Hypothesis}
We hypothesize that our method demonstrates comparable translation accuracy to existing baselines on a large and diverse data set of NL-TL pairs, with a strict guarantee on the validity of TL formulas, and running on a significantly smaller LLM (Gemma 3) which can be deployed onboard a robot with a single consumer GPU.
This experiment is designed to test our NL to TL approach in a broad, previously unseen domain.

\subsubsection*{Experiment Design}
To benchmark NL to TL translation performance, we use the dataset introduced by \cite{wang2021learning}, which consists of 7,500 natural language and linear temporal logic (LTL) pairs in the context of navigation tasks.

We compare our method to several ablated variants and to a GPT-4 baseline presented in \cite{chen2023nl2tl}.
Each method is evaluated with 5 and 20 few shot examples, with 500 NL-TL pairs randomly sampled from the dataset. Each evaluation is rerun 3 times, with resampled few-shot examples and NL-TL pairs.
In our approach, both the BNF grammar and few shot examples are included in the prompt, and the BNF grammar is enforced during decoding using \texttt{llama.cpp}'s constrained generation.
Results with constrained generation are not presented for GPT-4, as this capability does not exist for this LLM.
The ablations include variants without the grammar constraint, without the grammar prompt, and with combinations thereof, allowing us to isolate the contributions of each component of our method.

\subsubsection*{Metrics}
We measure Accuracy as percentage of generated LTL formulas logically equivalent to true LTL formulas from the dataset and Validity as percentage of generated formulas that are syntactically correct.
Logical equivalence and syntactic correctness of LTL formulas are checked using the \texttt{Spot} library \cite{duret2022spot}.

\subsubsection*{Results and Discussion}
The results are summarized in \Cref{tab:ablation}.
Our method achieves 100\% formula validity in all cases, confirming our hypothesized guarantee.
Additionally, we show similar translation accuracy in the unseen navigation dataset with a significantly smaller LLM.
Notably, including the BNF grammar in the prompt without the grammar constraint achieves higher accuracy than constrained generation, illustrating a drawback of this approach.
Addressing this issue is the subject of current works, which have proposed new constrained decoding algorithms \cite{beurer2024guiding}.
See \Cref{tab:full_ablation} for full results, including inference time and additional ablations.




\begin{table}[ht]
    \centering
    \resizebox{\columnwidth}{!}{%
    \begin{tabular}{clcc}
        \toprule
        \textbf{\# Ex.} & \textbf{Method Variant} & \textbf{Accuracy (\%)} & \textbf{Validity (\%)}\\
        \midrule
        \multirow{5}{*}{5} 
        & Gemma F + P + C (Ours)         & 56.80 $\pm$ 8.76  & \textbf{100.0} $\pm$ 0.0 \\
        & \phantom{Gemma} F + P             & \textbf{63.00} $\pm$ 5.87  & 99.8 $\pm$ 0.45  \\
        & \phantom{Gemma} F          & 56.80 $\pm$ 4.76  & 99.0 $\pm$ 1.22 \\
        & GPT-4 F + P             & 64.10 $\pm$ 8.79  & 100.0 $\pm$ 0.0 \\
        \midrule
        \multirow{6}{*}{20} 
        & Gemma F + P + C (Ours)         & 76.73 $\pm$ 2.93  & 100.0 $\pm$ 0.0 \\
        & \phantom{Gemma} F + P             & \textbf{89.73} $\pm$ 2.37  & \textbf{100.0} $\pm$ 0.0 \\
        & \phantom{Gemma} F          & 86.00 $\pm$ 0.40  & 100.0 $\pm$ 0.0 \\
        & GPT-4 F + P             & 92.73 $\pm$ 3.00  & 100.0 $\pm$ 0.0 \\
        & \phantom{GPT-4} F (baseline) & 87            & --\\
        \bottomrule
    \end{tabular}
    }
        \caption{Ablation study on NL to TL with varying number of few-shot examples. F, P, and C refer to few shot prompting, inclusion of the BNF grammar in the prompt, and the BNF grammar constrained generation respectively.}
        \vspace{-15pt}
    \label{tab:ablation}        
\end{table}


\subsection{Experiment 2: Mobile Robot Blocked by a Pallet}


Our second experiment is a case study in which forklift robots respond to a help request from a mobile robot prompted by the scene description $\confmsg =$ ``A pallet is blocking the entrance to the picking aisle.''
Each forklift robot $j \in \allhelpers$ then evaluates whether it can handle this additional task (moving the obstructing pallet) on top of its existing pick-and-place (PNP)\footnote{Our STL specification of a PNP job is in \Cref{appx: pick and place specification}.} jobs by updating its STL specification $\proposition\orig_j$. 
The help task, $\proposition_j\help$, is similarly encoded as a PNP job where the forklift must travel to the help site, pick up the obstructing pallet and place the pallet in the nearest free cell $\position\lbl{adj} = \arg\min_{\position}\{D(\position,\position\conf) \mid \position \in \freecells\}$.

\subsubsection*{Hypothesis}
We hypothesize that, by optimizing the helper's path to minimize both the requester's waiting time and the additional time incurred by the helper when providing assistance, we can significantly reduce the total extra time across all agents compared to common heuristic-based approaches.

\subsubsection*{Experiment Design}



We conducted 100 simulation trials with randomly chosen help-site locations.
In each trial, six forklift robots are spawned with random initial positions. 
Each forklift has two randomly assigned PNP tasks. 
Obstacles (including the blocking pallet) remain fixed in known locations.

We compare with two heuristic baselines: (H1) choosing the forklift robot closest to the help site, and (H2) selecting the first forklift that returns a feasible solution (i.e., the robot whose Gurobi solver finishes first).

\subsubsection*{Results and Discussion}

\begin{figure}
    \centering
    \includegraphics[width=7cm]{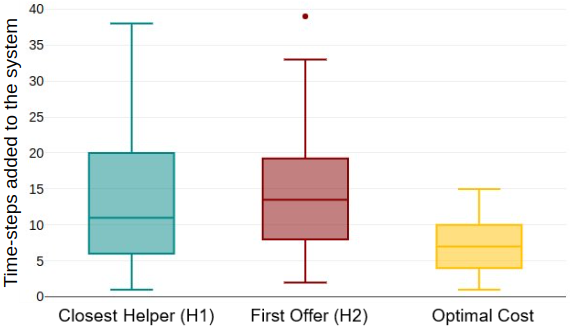}
    \caption{Comparison of total time-steps added to the system under heuristics and the optimal solution. 
    The closest helper (H1) coincided with the optimal solution in 42\% of trials while the first offer (H2) was optimal in only 25\% of trials.}
    \vspace{-15pt}
    \label{fig:box}
\end{figure}

\Cref{fig:box} summarizes our findings.
Our \emph{optimal cost} approach on average provides 26\% and 40\% efficiency gains over H1 and H2 respectively.
We observe that the interdependencies in PNP jobs make distance-based heuristics (H1) suboptimal, consistent with previous literature in operations research \cite{de2007design}.
Furthermore, due to the NP-Hardness of Mixed Integer Programs \cite{hartmanis1982computers}, faster compute time does not imply global optimality \cite{belta2019formal}.

%% file: sections/06_conclusion.tex
\section{Conclusion and Future Work}\label{sec: conclusion}
We have presented a novel framework that takes in scene descriptions from VLMs, identifies conflicts, and triggers help requests. Then, the help task is converted into a set of STL constraints, generated using a BNF grammar, which are solved as a MILP to minimize the additional time incurred by all agents.

We evaluated our approach in two scenarios. First, we benchmarked our method to translate natural language to TL on a large-scale navigation dataset, and showed comparable accuracy to existing methods with improved TL validity on small, local LLMs. Second, in a case study with forklifts and a blocked mobile robot, our method showed clear improvements compared to heuristic approaches by optimizing for total added time, resulting in more efficient system-level performance. 

For future work, we intend to explore scenarios where the VLM scene descriptions are uncertain or incomplete regarding the need for assistance. Furthermore, we plan to strengthen our framework with formal safety guarantees at runtime by considering methods such as Control Barrier Functions (CBFs).

%% file: sections/08_appendix.tex
\newpage

\section{Encoding STL Specifications}\label{appx: encoding stl}
We propose a lightweight interface to express navigational tasks as STL specifications.
Inspired by \texttt{stlpy} \cite{kurtz2022mixed}, our method emphasizes extracting \emph{spatial} propositions and decoupling \emph{temporal} constraints.
We detail our approach in the appendix, and provide an illustrative example here.

Consider encoding the sentence ``\textit{Visit Aisle 1 at least once before time $T$ while avoiding obstacles},''.
The corresponding STL expression is:
\begin{align}
    \proposition\lbl{spec} = \always_{[0,\timehorizon]} \lnot \: \regtext{Obstacle} \land \eventually _{[0,T]}\: \regtext{Aisle1}
    \label{eq: stlpy vs our stl example}
\end{align}
where Aisle1 and Obstacle are atomic propositions (AP) tied to the corresponding coordinates in the environment, and become true whenever the agent is inside the region.

We then specify the time horizon $T$ independently when setting up the optimization problem. The resulting code snippet is:
 \begin{mycodelisting}
model.spec = F(aisle1) & G(NOT(obstacle))
model.T = T
\end{mycodelisting}
\subsection{Global STL Specification}\label{appx: global stl spec}

Each robot in our framework is subject to a global specification of safety and actuation constraints.
For example, all ground vehicles must avoid obstacle cells and can only move one grid space in any of the cardinal directions.
Therefore all ground robots in the system are subjected to:
\begin{equation}
\proposition\lbl{global} = \always(\lnot \regtext{Obstacle} \land \mathcal{L}\lbl{actuation})
\end{equation}
\vspace{-20pt}
\subsection{Pick-and-Place (PNP) Tasks as STL Specifications}\label{appx: pick and place specification}
We create a PNP specification as
\begin{align*}
    &\varphi_{1} = (\lnot \position\lbl{place} \; \textbf{U}_{[0,\timehorizon]} \:\position\lbl{pick})\\
    &\textit{(Robot must pickup the pallet before placing it)}\\
    &\varphi_{2} = \Box_{[0,\timehorizon]}(\position\lbl{pick}\rightarrow(\lnot \position\lbl{others} \textbf{U}_{[0,\timehorizon]}\position\lbl{place})\\
    &\textit{(Pallet picked up must be placed before starting other tasks)}\\
    &\varphi_{3} = \Diamond_{[0,\timehorizon]} \position\lbl{place}\\
    &\textit{(Pallet is eventually placed within time horizon $\timehorizon$)}\\
    &\varphi_{\text{pnp}}(\position\lbl{place}, \position\lbl{pick}) = \varphi_1 \land \varphi_2 \land \varphi_3 
\end{align*}
\subsection{Example LTL BNF Grammar}\label{appx:bnf_example} 
\begin{mycodelisting}
root ::= ltl-depth-2
ltl-depth-2 ::= proposition 
             | "~(" ltl-depth-1 ")"
             | ltl-depth-1 binary-op ltl-depth-1 
             | unary-op "(" ltl-depth-1 ")"

ltl-depth-1 ::= proposition 
             | "~(" proposition ")" 
             | proposition binary-op proposition 
             | unary-op "(" proposition ")"

proposition ::= 'go_to_flag' | 'pass_by_trash_bin'

binary-op ::= " & " | " | " | " -> " | " U "
# '&' (and): both propositions must be true
# '|' (or): at least one predicate must be true
# '->' (implies): if first predicate is true, then second predicate must be true 
# 'U' (until): first predicate must be true at least until second predicate is true

unary-op ::= globally | eventually
globally ::= "G"
# Predicate must always be true at every timestep

eventually ::= "F"
# Predicate must be true at some time in the future
\end{mycodelisting}

\newpage
\section{Natural Language}
\subsection{Example Help Request}\label{appx:example_request}

\begin{quote}
    \ttfamily
    \small
"A pallet is blocking the aisle at location (1, 4). Assistance is required to move the pallet. Capabilities needed: 'lift' or 'push\_heavy\_objects' to clear the obstruction."  

Reasoning:\\  
1. Why I can’t resolve this alone: My "move" capability only allows navigation, not manipulating objects like pallets.  \\
2. Required capabilities: A robot with "lift" or "push\_heavy\_objects" is needed to physically relocate the pallet. \\ 
3. Request clarity: Specifying the obstacle type (pallet), location (1, 4), and required capabilities ensures helpers can assess their ability to assist.
\end{quote}

\subsection{Example Help Offer}\label{appx:example_proposal}
\begin{quote}
\ttfamily
\small
Assistance Proposal:\\
1. Capabilities Utilized:\\
   - Lift Pallet: The robot can securely lift and relocate the pallet.\\
   - Move: The robot can navigate to the target location and transport the pallet.\\  

2. Effort Analysis:  \\
   - Task Complexity: Low to moderate. Moving a single pallet is within standard operational parameters.  \\
   - Navigation: Direct path to (4, 0) is assumed clear (no additional obstacles mentioned).  \\

Location: (4, 0)  \\
The robot will move to the specified location, lift the pallet, and relocate it to an unoccupied storage zone to clear the aisle.    
\end{quote}

\section{Experiment I Full Results}\label{appx: exp 1 full results}

Full results of the ablation study on NL to TL with varying number of few-shot examples are presented in \Cref{tab:full_ablation}.
F, P, and C refer to few shot prompting, inclusion of the BNF grammar in the prompt, and the BNF grammar constrained generation respectively.
For GPT-4, inference time is the time from an API call to the model to a response, which includes network latency in addition to LLM inference time.
\begin{table}[ht]
    \centering
    \resizebox{0.7\columnwidth}{!}{%
    \begin{tabular}{llccc}
        \toprule
        \textbf{\# Examples} & \textbf{Method Variant} & \textbf{Accuracy (\%)} & \textbf{Validity (\%)} & \textbf{Inference Time (ms)} \\
        \midrule
        \multirow{5}{*}{5} 
        & Gemma F + P + C (Ours)         & 56.80 $\pm$ 8.76  & 100.0 $\pm$ 0.0   & 801.6 $\pm$ 36.2 \\
        & \phantom{Gemma} F + P             & 63.00 $\pm$ 5.87  & 99.8 $\pm$ 0.45   & 583.4 $\pm$ 28.2 \\
        & \phantom{Gemma} F          & 56.80 $\pm$ 4.76  & 99.0 $\pm$ 1.22   & 410.2 $\pm$ 9.2 \\
        & \phantom{Gemma} F + C             & 49.40 $\pm$ 5.59  & 100.0 $\pm$ 0.0   & 629.4 $\pm$ 26.1 \\
        & GPT-4 prompt, no constraint             & 64.10 $\pm$ 8.79  & 100.0 $\pm$ 0.0   & 803.7 $\pm$ 30.3 \\
        \midrule
        \multirow{6}{*}{20} 
        & Gemma F + P + C (Ours)         & 76.73 $\pm$ 2.93  & 100.0 $\pm$ 0.0   & 1191.7 $\pm$ 14.1 \\
        & \phantom{Gemma} F + P             & 89.73 $\pm$ 2.37  & 100.0 $\pm$ 0.0   & 974.9 $\pm$ 2.5 \\
        & \phantom{Gemma} F          & 86.00 $\pm$ 0.40  & 100.0 $\pm$ 0.0   & 414.9 $\pm$ 3.9 \\
        & \phantom{Gemma} F + C             & 72.27 $\pm$ 2.89  & 100.0 $\pm$ 0.0   & 633.5 $\pm$ 8.5 \\
        & GPT-4 F + P             & 92.73 $\pm$ 3.00  & 100.0 $\pm$ 0.0   & 2736.9 $\pm$ 25.2 \\
        & \phantom{GPT-4} F (baseline) & 87            & --            & -- \\
        \bottomrule
    \end{tabular}
    }
    \label{tab:full_ablation}        
\end{table}

\newpage

\section{Related Work} \label{appx: related work}

This work lies at the intersection of foundation models and formal methods for multi-agent robot task planning.

\subsection{Foundation Models in Multi-Robot Settings}

Foundation models have been specifically adapted for failure resolution in multi-robot systems (MRS).
Closest to our work, \cite{kato2024design} proposes a heterogeneous MRS where each robot uses a VLM to detect and describe impediments to task progress and an LLM to reason over robot capabilities to determine when to ask for and offer help.
However, this approach relies on foundation models for reasoning, providing no safety or optimality guarantees; we adopt a similar framework but leverage formal logic to address these limitations.
Conflicts between robots, as opposed to conflicts with the environment, can also be resolved using foundation models.
For example, VLMs can resolve deadlock in a connected MRS by choosing a leader agent and a set of waypoints for the leader using descriptions of the task and environment \cite{garg2024foundation}; note that our work focuses on environment conflicts.

Finally, foundation model-driven MRS are typically assessed on success rates of various tasks, but have not been studied extensively in terms of efficiency or optimality.
An attempt in this direction uses conformal prediction on an LLM's built-in confidence score to determine whether multi-robot task and motion plans are probabilistically correct \cite{wang2024probabilistically}.
In contrast, our work uses formal logic and a convex solver to enable a foundation model to assess optimality.

\subsection{Formal Methods in Robotics}

Formal methods, such as Temporal Logic (TL), offer robust guarantees for MRS task specifications, but often assume static environments and specifications \cite{ulusoy2013optimality}.
Recent literature attempts to tackle this limitation by making LTL planning more robust and dynamic.
For example, one can use multi-robot and human-in-the-loop (HIL) planning where robots share new conflicts and synthesize alternate trajectories to satisfy their LTL specifications \cite{yu2023reactive}.
Similarly, but in a single-agent scenario, \cite{ren2024ltl} takes an incremental optimal replanning strategy for both cases where the satisfaction of the LTL specification is feasible and infeasible.
Finally, \cite{liu2024lang2ltl} has the closest resemblance to our work, using a VLM to ground spatio-temporal navigation commands in unseen environments for a single agent case. In contrast to the previous works, we propose a novel multi-agent framework that integrates STL to augment an LLM agent with spatial and temporal reasoning capabilities.